\documentclass[10pt,twocolumn,letterpaper]{article}

\usepackage{iccv}
\usepackage{times}
\usepackage{epsfig}
\usepackage{graphicx}
\usepackage{amsmath}
\usepackage{amssymb}
\usepackage{bbm}
\usepackage{subcaption}
\graphicspath{{figures/}}

\newtheorem{theorem}{Theorem}

\usepackage[pagebackref=true,breaklinks=true,letterpaper=true,colorlinks,bookmarks=false]{hyperref}

\iccvfinalcopy

\def\iccvPaperID{148}

\begin{document}

\title{Analytical Moment Regularizer for Gaussian Robust Networks}

\author{Modar Alfadly\thanks{Both authors contributed equally to this work.}, Adel Bibi\footnotemark[1], and Bernard Ghanem\\
King Abdullah University of Science and Technology (KAUST), Saudi Arabia\\
{\tt\small \{modar.alfadly,adel.bibi,bernard.ghanem\}@kaust.edu.sa}}
\maketitle

\begin{abstract}
\label{abstract}
Despite the impressive performance of deep neural networks (DNNs) on numerous vision tasks, they still exhibit yet-to-understand uncouth behaviours. One puzzling behaviour is the subtle sensitive reaction of DNNs to various noise attacks. Such a nuisance has strengthened the line of research around developing and training noise-robust networks. In this work, we propose a new training regularizer that aims to minimize the probabilistic expected training loss of a DNN subject to a generic Gaussian input. We provide an efficient and simple approach to approximate such a regularizer for arbitrary deep networks. This is done by leveraging the analytic expression of the output mean of a shallow neural network; avoiding the need for the memory and computationally expensive data augmentation. We conduct extensive experiments on LeNet and AlexNet on various datasets including MNIST, CIFAR10, and CIFAR100 demonstrating the effectiveness of our proposed regularizer. In particular, we show that networks that are trained with the proposed regularizer benefit from a boost in robustness equivalent to performing 3-21 folds of data augmentation.

\end{abstract}

\vspace{-20pt}
\section{Introduction}
\label{introduction}

\begin{figure}[t]
    \centering
    \includegraphics[width=0.45\textwidth]{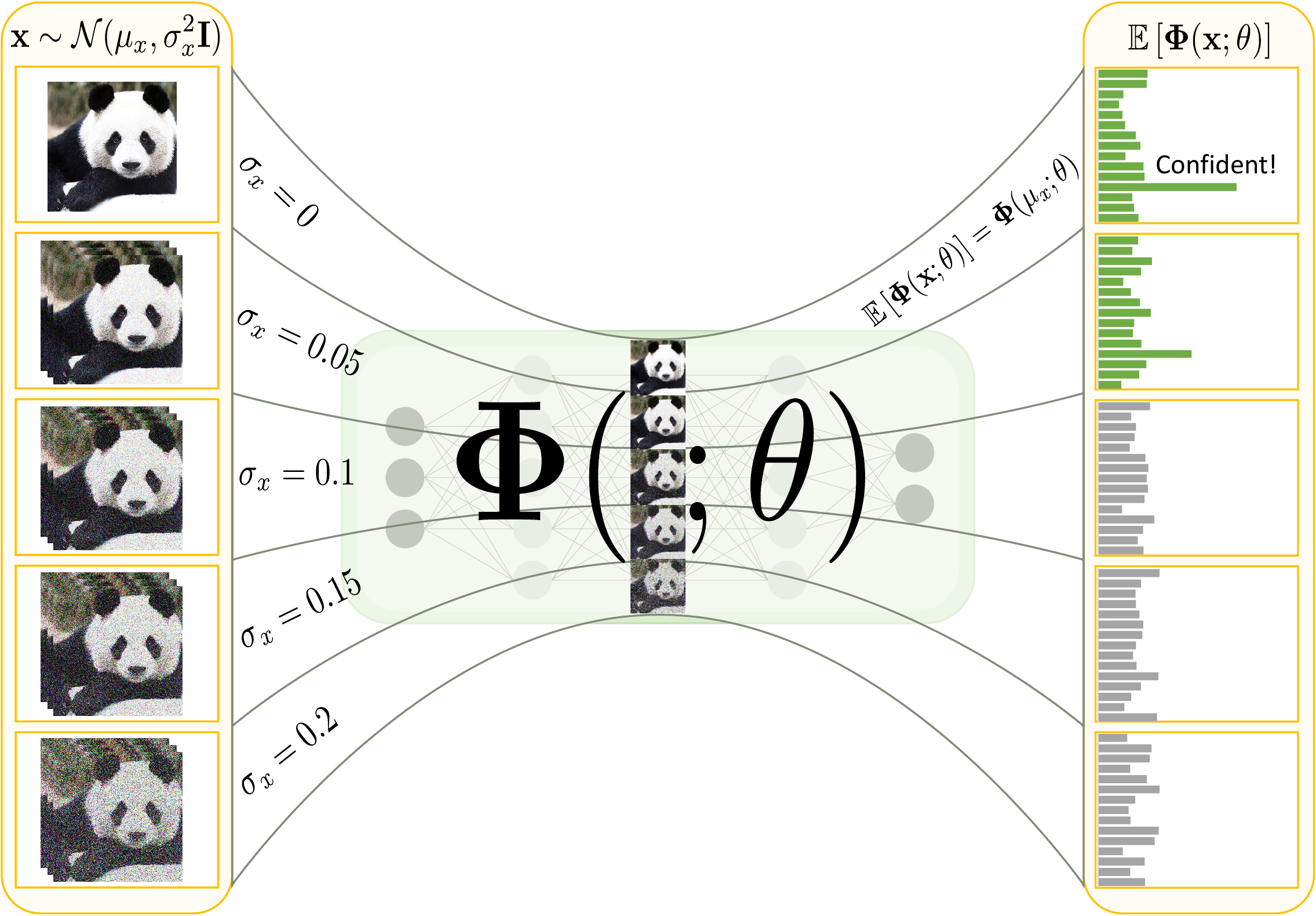}
    \caption{\textbf{The confidence of a classifier is negatively affected by Gaussian input noise.} as we increase the level of noise $\sigma_x$ (i.e., on the left from top to bottom), the confidence of the classifier decreases; see the mean output logits on the right from top to bottom. However, this is expected to happen but the rate in which the confidence decays is the factor that decides the robustness of the classifier. It can be slowed down through computationally expensive data augmentation or using our proposed lightweight regularizer.}
    \label{fig:illustrative_fig}
    \vspace{-20pt}
\end{figure}

Deep neural networks (DNNs) have emerged as generic models that can be trained to perform impressively well in both vision related tasks (\eg object recognition \cite{he2016deep} and semantic segmentation \cite{long2015fully}) and non-vision related tasks (\eg speech recognition \cite{hinton2012deep} and bio-informatics \cite{angermueller2016deep}). Despite their increasing popularity, flexibility, generality, and performance, DNNs have been recently shown to be quite susceptible to small imperceptible input noise \cite{szegedy2013intriguing,moosavi2016deepfool,goodfellow2014explaining}. Such analysis gives a clear indication that even state-of-the-art DNNs may lack robustness. Consequently, there has been an ever-growing interest in the computer vision and machine learning communities to study this uncanny behaviour. In particular, the work of \cite{moosavi2016deepfool,goodfellow2014explaining} demonstrates that there are systematic approaches to constructing adversarial attacks that result in misclassification errors with high probability. Even more peculiarly, some noise perturbations seem to be doubly agnostic \cite{moosavi2017universal}, \ie there exists a deterministic perturbation that can result in misclassification errors with high probability when applied to different networks irrespective of the input image (network and input agnostic).

Understanding this degradation in performance under adversarial attacks is of tremendous importance, especially for real-world DNN deployments (\eg, self-driving cars/drones and equipment for the visually impaired). A standard and popular mean to alleviate this nuisance is noisy data augmentation in training, \ie a DNN is exposed to noisy input images during training so as to bolster its robustness during inference. Several works have demonstrated that DNNs can in fact benefit from such augmentation \cite{moosavi2016deepfool,goodfellow2014explaining}. However, data augmentation in general might not be sufficient for two reasons. (1) Particularly with high-dimensional input noise, the amount of data augmentation necessary to sufficiently capture the noise space will be very large, which will increase the training time. (2) Data augmentation with high energy noise can negatively impact the performance on noise-free test examples. This can be explained by the fundamental trade-off between accuracy and robustness \cite{tsipras2018robustness,Boopathy2019cnncert}. It can also arise from the fact that the augmentation forces the DNN to have the same prediction for two vastly different versions of the same input: noise-free and a substantially corrupted version. Therefore, we propose a new regularizer for noise-robust networks to circumvent the aforementioned setbacks of data augmentation.

A natural objective to use for training against attacks sampled from a distribution $\mathcal{D}$, that bypasses the need of data augmentation, is the \emph{expected loss} under such distribution. Since a closed-form expression is generally difficult to obtain or an approximate surrogate is expensive to evaluate (Monte Carlo estimates), we propose instead a closely related objective that is the loss of the \emph{expected predictions} of the network under $\mathcal{D}$-distributed adversarial noise. Since it has been shown that Gaussian noise can be adversarial \cite{Bibi_2018_CVPR} and that such noise is widely studied in image processing applications, we restrict the work in this paper to the case where $\mathcal{D}$ is Gaussian. However, even under such an assumption, only a notoriously memory and computationally expensive closed-form approximate surrogate for the \emph{expected predictions} exist \cite{Bibi_2018_CVPR} through means of performing a two-stage linearization to the network. To this end, we propose a new approach that allows for simple and efficient evaluation of our proposed regularizer (the \emph{expected prediction} loss), surpassing formidable network linearizations and prohibitive data augmentation. Such an approach will decrease training cost, as it would replace commonly used and generally effective data augmentation strategies that requires a large number of noisy training examples (especially as the dimensionality of the problem increases \cite{tramer2017space}). As a result, while being faster, we can achieve similar or mostly better robustness to 3-21 folds of noisy data augmentation.

\vspace{-10pt}
\paragraph{\textbf{Contributions.}} \textbf{(i)} We formalize a new regularizer that is a function of the probabilistic first moment of the output of a DNN, to train robust DNNs against noise sampled from distribution $\mathcal{D}$. \textbf{(ii)} Under the special choice of Gaussian attacks, \ie $\mathcal{D}$ is Gaussian, we show how the first moment expression can be evaluated very efficiently during training for an arbitrary deep DNN by bypassing the need to perform memory and computationally expensive two-stage linearization. \textbf{(iii)} Extensive experiments on various architectures (LeNet \cite{lecun1999object} and AlexNet \cite{krizhevsky2012imagenet}) and on several datasets (MNIST \cite{lecun1998mnist}, CIFAR10, and CIFAR100 \cite{krizhevsky2009learning}) demonstrate that a substantial enhancement in robustness can be achieved when using our regularizer in training. In fact, in the majority of the experiments, the improvement is better than training on the same dataset but augmented with 3 to 21 times noisy data. Interestingly, the results suggest an excellent trade-off between accuracy and robustness.

\section{Related Work}
\label{related_work}
Despite the impressive performance of DNNs on various tasks, they have been proven to be very sensitive to certain types of noise, commonly referred to as adversarial examples, particularly in the recognition task \cite{moosavi2016deepfool,goodfellow2014explaining}. Adversarial examples can be viewed as small imperceptible noise that once added to the input of a DNN, the performance is severely degraded. This finding has broadened the interest in studying and measuring the robustness of DNNs. 

The work of \cite{szegedy2013intriguing} suggested a spectral stability analysis for a wide class of DNNs by measuring the Lipschitz constant of the affine transformation describing a fully-connected or a convolutional layer. This result was extended to compute an upper bound for a composition of layers (\ie, a DNN). However, this measure only sets an upper bound on the robustness over the \emph{entire} input domain and does not take into account the noise distribution. A later work defined robustness as the mean support of the minimum adversarial perturbation \cite{fawzi2017robustness}, which is the most common definition for robustness. It studied robustness against not only adversarial perturbations but also against geometric transformations to the input. The authors of  \cite{Fawzi2018} emphasized the independence of the robustness measure to the ground truth class labels and that it should only depend on the classifier and the dataset distribution. Subsequently, they proposed two different metrics to measure DNN robustness: one for general adversarial attacks and the other for noise sampled from a uniform distribution. In \cite{Jean2018}, theoretical bounds on the robustness of linear classifiers to Gaussian and uniform noise were derived as a function of the support of the minimum adversarial perturbation. This work was then extended to DNNs with locally \emph{approximately flat} decision boundaries, which is a common phenomenon in state-of-the-art DNNs. Very recently, the work of \cite{gilmer2018adversarial} showed the trade-off between robustness and test error from a theoretical point of view on a simplified classification problem with hyper-spheres.

On the other hand, and based on various robustness analyses, several works proposed various approaches in building networks that are robust against noise sampled from well known distributions and against generic adversarial attacks. For instance, the work of \cite{grosse2017statistical} proposed a model that was trained to classify adversarial examples with statistical hypothesis testing on the distribution of the dataset. Another approach is to perform statistical analysis on the latent feature space \cite{li2016adversarial,feinman2017detecting} or train a DNN that rejects adversarial attacks \cite{lu2017safetynet}. Moreover,  the geometry of the decision boundaries of DNN classifiers was studied in \cite{Fawzi2017ClassificationRO}  to infer a simple curvature test for this purpose. Using this method, one can restore the original label and classify the input correctly. Restoring the original image/label using defense mechanisms, which can only detect adversarial examples, can be done by denoising the image (ridding it from its adversarial nature) so long as the noise perturbation is well-known and modeled apriori \cite{Zhu_2016_CVPR}. Motivated by the same goal, the authors of \cite{das2017keeping,dziugaite2016study} studied the effect of JPEG image compression on denoising adversarial examples. A more involved solution would be to build models that are intrinsically more robust to this type of noise \cite{gu2014towards}. For example, the use of bounded ReLU was suggested in \cite{zantedeschi2017efficient} to encourage robustness by limiting the output range. They complemented this model with data augmentation of Gaussian noise, as it has been shown to be a direct approach in improving robustness. However, this can be far less effective as the dimensionality of the problem increases. A different work proposed to distill the learned knowledge from a deep model to retrain a similar model architecture as means to improving robustness \cite{papernot2016distillation}. This training approach is one of many adversarial training strategies for robustness \cite{makhzani2015adversarial}. More closely to our work is \cite{cisse2017parseval}, where  a new training regularizer was proposed for a large family of DNNs. The proposed regularizer softly enforce that the upper bound of the Lipshitz constant of the output of the network to be less than or equal to one. Moreover and very recently, the work of \cite{Bibi_2018_CVPR} has derived analytic expressions for the output mean and covariance of networks in the form of (Affine, ReLU, Affine) under a generic Gaussian input. This work also demonstrates how a (memory and computation expensive) two-stage linearization can be employed to locally approximate a deep network with a two layer version; thus, enabling the application of the derived expressions on the approximate shallow network. 

We will first start with an overview of the Gaussian mean expression from \cite{Bibi_2018_CVPR}. Next, we propose our new regularizer that is the loss of the expected network prediction under input noise sampled from general distribution $\mathcal{D}$ and how it relates to data augmentation approaches. Then, we propose our efficient lightweight approach to evaluate the approximate surrogate of the expected output predictions without neither performing expensive network linearizations nor data augmentation when $\mathcal{D}$ is Gaussian. Lastly, we present extensive experimental results verifying our contributions.

\section{Methodology}
\label{methodology}

\begin{figure*}[t]
    \centering
    \includegraphics[width=0.95\textwidth]{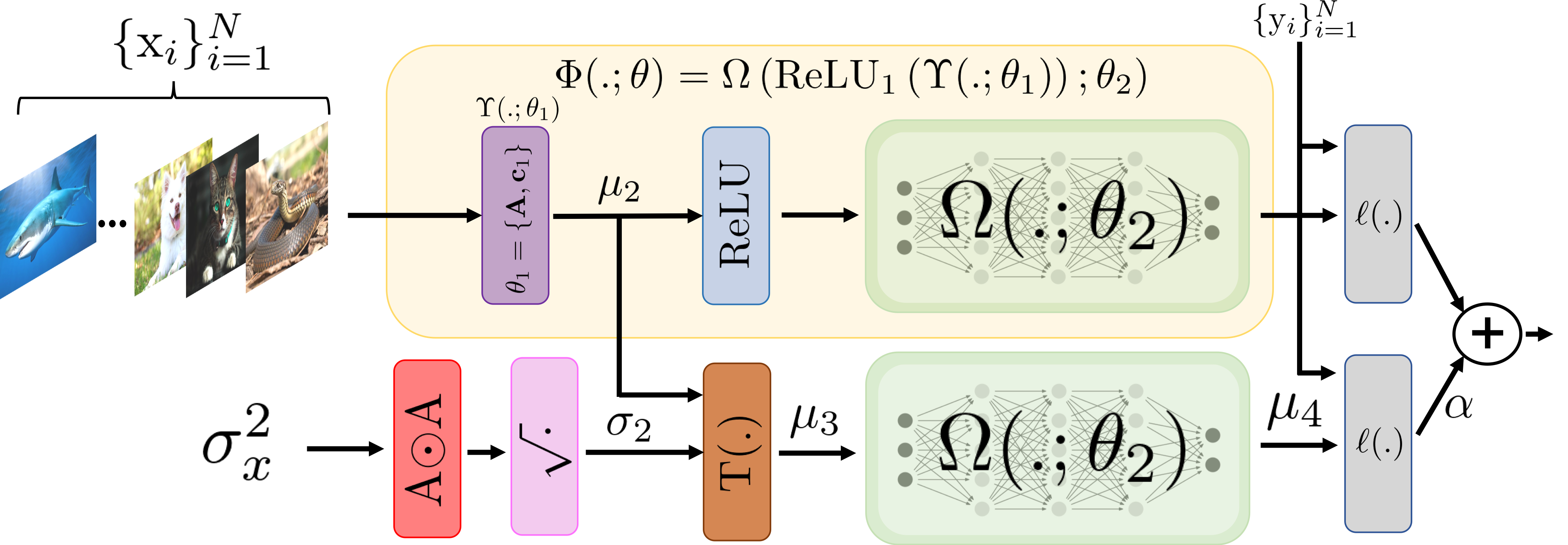}
    \caption{\textbf{Overview of the proposed graph for training Gaussian robust networks} using very efficient approximation to the proposed regularizer in Eq \eqref{objective_all}. The yellow block corresponds to an arbitrary network $\Phi(.,\theta)$ viewed as the composition of two subnetworks separated by a $\text{ReLU}$ as nonlinearity. The stream on the bottom computes the output mean $\mu_4$ of the network $\Phi(:,\theta)$ under the assumption that (i) the noise input distribution is independent Gaussian with variances $\sigma_x^2$ (ii) $\Omega(.:\theta_2)$ is approximated by a linear function. Such evaluation for the output mean is efficient as it requires an extra forward pass as opposed to any form of computationally and memory intensive network linearizations or data augmentation.}
    \label{fig:pipeline_fig}
\end{figure*}

\subsection{Background on Network Moments}
Networks with a single hidden layer of the form (Affine, ReLU, Affine) can be written in the functional form $\mathbf{g}(\mathbf{x}) = \mathbf{B} \text{max} \left(\mathbf{A} \mathbf{x} + \mathbf{c}_1,\mathbf{0}_p \right) + \mathbf{c}_2$. The $\text{max}(.)$ is an element-wise operator, $\mathbf{A} \in \mathbb{R}^{p \times n}$, and $\mathbf{B} \in \mathbb{R}^{d \times p}$. Thus, $\mathbf{g} : \mathbb{R}^{n} \rightarrow \mathbb{R}^{d}$. 
Given that $\mathbf{x} \sim \mathcal{N} \left(\mu_x,\Sigma_x\right)$, Bibi \etal \cite{Bibi_2018_CVPR} showed that\footnote{This is the more compact expression in the author's project page.}:
\begin{theorem} The first moment of $g(\mathbf{x})$ is:
    \label{theorem_1}
    \begin{equation}
        \begin{aligned}
        \resizebox{0.90\hsize}{!}{
       $\mathbb{E}[\mathbf{g}(\mathbf{x})] = \mathbf{B} \underbrace{\left(\mu_2 \odot \Phi\left(\frac{\mu_2}{\sigma_2}\right) + \sigma_2 \odot\varphi \left(\frac{\mu_2}{\sigma_2}\right)\right)}_{\mu_3 = T(\mu_2,\sigma_2)} + \mathbf{c}_2$}
        \end{aligned}
    \end{equation}
\end{theorem}
\noindent where $\mu_2 = \mathbf{A}\mu_x+\mathbf{c}_1$, $\sigma_2 = \sqrt{\text{diag}\left(\Sigma_2\right)}$, $\mathbf{\Sigma}_2 = \mathbf{A}\Sigma_x\mathbf{A}^\top$, $\Phi$ and $\varphi$ are the standard Gaussian cumulative (CDF) and density (PDF) functions, respectively. The multiplication $\odot$ and division are element-wise operators. Lastly, $\text{diag}(.)$ extracts the diagonal elements of a matrix into a vector.

To extend the results of Theorem (\ref{theorem_1}) to deeper models,  a two-stage linearization was proposed in \cite{Bibi_2018_CVPR}, where ($\mathbf{A}$, $\mathbf{B}$) and ($\mathbf{c}_1$, $\mathbf{c}_2$) are taken to be the Jacobians and biases of the first order Taylor approximation to both functions around a certain ReLU layer in a DNN. Please, refer to \cite{Bibi_2018_CVPR} for more details about this expression and the proposed linearization.

\subsection{Proposed Robust Training Regularizer}
Training DNNs that are robust to certain types of noise, \eg noise examples from distribution $\mathcal{D}$, is still an elusive problem. The most standard straightforward approach that is often used in practice is to sample noise from $\mathcal{D}$ and train the network with the noise-augmented dataset. However, while this approach is in general effective, it comes with two major complications. First, if the input space (size of input data) is large, the amount of noisy examples to be augmented to the dataset might necessarily be large for it to be effective. Thus, training time will be severely impacted, and as a consequence, robustness to different noise levels will be infeasible. Second, if the noise level of the augmented data is very high, the performance of the network on noise-free examples will also be degraded, since the network is presented during training with two very different examples, noise-free and a heavily corrupted counterpart, both of which have the same class label.
    
To propose an alternative approach to noisy data augmentation that does not suffer from its drawbacks, one has to realize that this augmentation strategy aims to minimize the expected training loss of a DNN when subjected to a noisy input distribution $\mathcal{D}$ through sampling. In fact, it minimizes an empirical loss that approximates this expected loss when enough samples are presented during training. When sampling is insufficient (a drawback of data augmentation in high-dimensions), this approximation is too loose and robustness can suffer.  However, if we have access to an analytic expression for the expected loss of the network, expensive data augmentation and sampling can be averted. This is the key motivation of the paper. Mathematically, the training loss can be modeled as:

\begin{equation}
  \setlength{\abovedisplayskip}{-5pt} \setlength{\abovedisplayshortskip}{-5pt}
    \label{objective_all_or}
    \begin{aligned}
    \min_{\theta} ~ \sum_{i=1}^N \Big( \ell\left(\Phi(\mathbf{x}_i;\theta), y_i \right) + \alpha\mathbb{E}_{\mathbf{n}\sim \mathcal{D}}\left[ \ell\left(\Phi(\mathbf{x}_i + \mathbf{n};\theta), y_i \right)\right]\Big)
    \end{aligned}
\end{equation}

\noindent Here, $\Phi: \mathbb{R}^{n} \rightarrow \mathbb{R}^d$ is any arbitrary network with parameters $\theta$, $\ell$ is the loss function, $\{(\mathbf{x}_i,y_i)\}_{i=1}^N$ are the noise-free data-label training pairs and $\alpha \ge 0$ is a trade off parameter. While the first term in \eqref{objective_all_or} is the standard empirical loss commonly used for training, the second term is often replaced with its Monte Carlo estimates through data augmentation. That is, for each training example $\mathbf{x}_i$, the second term in \eqref{objective_all_or} is approximated with an empirical average of $\tilde{N}$ noisy examples of $\mathbf{x}_i$ such that $\mathbb{E}_{\mathbf{n}\sim \mathcal{D}}[\ell\left(\Phi(\mathbf{x}_i + \mathbf{n};\theta), y_i \right)] \approx  \frac{1}{\tilde{N}}\sum_{j=1}^{\tilde{N}} \ell\left(\Phi(\mathbf{x}_i + \mathbf{n}_j;\theta), y_i \right)$. This will increase the size of the dataset by a factor of $\tilde{N}$ (there are $\tilde{N}$ samples for each $\mathbf{x}_i$), which will in turn increase training complexity. As discussed earlier, network performance on the noise-free examples can also be negatively impacted. Note that obtaining a closed form expression for the second term in \eqref{objective_all_or} for some of the popularly used losses $\ell$ is more complicated than deriving expressions for the output mean of the network $\Phi$ itself, \eg in Theorem (\ref{theorem_1}). Therefore, we propose to replace this loss \eqref{objective_all_or} with the following surrogate:

\begin{equation}
  \setlength{\abovedisplayskip}{-5pt} \setlength{\abovedisplayshortskip}{-5pt}
    \label{objective_all}
    \begin{aligned}
    \min_{\theta} ~~ \sum_{i=1}^N \Big( \ell\left(\Phi(\mathbf{x}_i;\theta), y_i \right) + \alpha\ell\left(\mathbb{E}_{\mathbf{n}\sim \mathcal{D}}\left[\Phi(\mathbf{x}_i + \mathbf{n};\theta)\right], y_i\right) \Big)
    \end{aligned}
\end{equation}

\noindent Because of Jensen's inequality, Eq \eqref{objective_all} is a lower bound to Eq \eqref{objective_all_or} when $\ell$ is convex, which is the case for most popular losses including $\ell_2$-loss and cross-entropy loss. The proposed second term in \eqref{objective_all} encourages that the output mean of the network $\Phi$ of every noisy example $\mathbf{x}_i + \mathbf{n}$ matches the correct class label $y_i$. This regularizer will stimulate a separation among the output mean of the classes if the training data is subjected to noise sampled from $\mathcal{D}$. Having access to an analytic expression for these means will prompt a simple inexpensive training, where the actual size of the training set is unaffected and augmentation is avoided. This form of regularization is proposed to replace data augmentation.

While a closed-form expression for the second term of \eqref{objective_all} might be infeasible for a general network $\Phi(.)$, an expensive approximation can be attained. In particular, Theorem \eqref{theorem_1} provides an analytic expression to evaluating the second term in \eqref{objective_all}, for when $\mathcal{D}$ is Gaussian and when the network is approximated by a two-stage linearization procedure as $\Phi(\mathbf{x}) \approx \mathbf{B}\text{max}\left(\mathbf{A}\mathbf{x} + \mathbf{c}_1, \mathbf{0}_p\right) + \mathbf{c}_2$. However, it is not clear how to utilize such a result to regularize networks during training with \eqref{objective_all} as a loss. This is primarily due to the computationally expensive and memory intensive network linearization, \ie two-stage linearization proposed in \cite{Bibi_2018_CVPR}, which makes it challenging to expose the network parameters $\theta$ for efficient update in training \cite{thesis}. Specifically, the linearization parameters $(\mathbf{A},\mathbf{B},\mathbf{c}_1,\mathbf{c}_2)$ are a function of the network parameters, $\theta$, that are updated with every gradient descent step on \eqref{objective_all}; thus, two-stage linearization has to be performed in every $\theta$ update step, which is infeasible.
    
\subsection{On an Efficient Approximation to \eqref{objective_all}}
The loss in \eqref{objective_all} proposes a generic approach to train robust arbitrary networks against noise sampled from an arbitrary distribution $\mathcal{D}$. Since the problem in its general setting is too broad for detailed analysis, we restrict the scope of this work to the class of networks, which are most popularly used and parameterized by $\theta$, $\Phi(.;\theta) : \mathbb{R}^n \rightarrow \mathbb{R}^d$ with ReLUs as nonlinear activations. Moreover, since random Gaussian noise was shown to exhibit an adversarial nature \cite{Bibi_2018_CVPR,rauber2017foolbox,Jean2018}, and that it is one of the most well studied noise models in computer vision for the nice properties it exhibits, we restrict $\mathcal{D}$ to the case of Gaussian noise. In particular, $\mathcal{D}$ is independent zero-mean Gaussian noise at the input, \ie $\mathbf{n} \sim \mathcal{D} = \mathcal{N}\left(\mathbf{0},\Sigma_x = \text{Diag}\left(\sigma_x^2\right) \right)$, where $\sigma_x^2 \in \mathbb{R}^n$ is a vector of variances and $\text{Diag}(.)$ reshapes the vector elements into a diagonal matrix. Generally, it is still difficult to compute the second term in \eqref{objective_all} under Gaussian noise for arbitrary networks. However, if we have access to an inexpensive approximation of the network, avoiding the computationally and memory expensive network linearization in \cite{Bibi_2018_CVPR}, $\Phi$ in the form $\Phi(\mathbf{x}) \approx \mathbf{B}\text{max}\left(\mathbf{A}\mathbf{x} + \mathbf{c}_1, \mathbf{0}_p\right) + \mathbf{c}_2$, an approximation to the second term in \eqref{objective_all} can thus be used for efficient robust training directly on $\theta$.

Consider the $l^{\text{th}}$ ReLU layer in $\Phi(.;\theta)$; the network can be expressed as $\Phi(.;\theta) = \Omega(\text{ReLU}_l(\Upsilon(.,\theta_1));\theta_2)$. Note that the parameters of the overall network $\Phi(.;\theta)$ is the union of the parameters of the two subnetworks $\Upsilon(.;\theta_1)$ and $\Omega(.;\theta_2)$, \ie $\theta = \theta_1 \cup \theta_2$. Throughout this work and to simplify the analysis, we set  $l=1$. That is, under such choice of $l$, the first subnetwork $\Upsilon(.,\theta_1)$ is linear with $\theta_1 = \{\mathbf{A},\mathbf{c}_1\}$. However, the second subnetwork $\Omega(.,\theta_2)$ is not linear in general, and thus, one can linearize $\Omega(.,\theta_2)$ at $\mathbb{E}_{\mathbf{n} \sim \mathcal{N}(\mathbf{0},\Sigma_x)}\left[\text{ReLU}_1 \left( \Upsilon\left(\mathbf{x}_i + \mathbf{n};\theta_1\right)\right)\right] = T(\mu_2, \sigma_2) = \mu_3$. Note that $\mu_3$ is the output mean after the $\text{ReLU}$ and $\mu_2 = \mathbf{A}\mathbf{x}_i + \mathbf{c}_1$, since $\Upsilon(\mathbf{x}_i + \mathbf{n};\theta_1) = \mathbf{A}\left(\mathbf{x}_i + \mathbf{n}\right) + \mathbf{c}_1$. Both $T(.,.)$ and $\sigma_2$ are defined in \eqref{theorem_1}. Thus, linearizing $\Omega$ at $\mu_3$ with linearization parameters $\{\mathbf{B},\mathbf{c}_2\}$, where $\mathbf{B}$ is the Jacobian of $\Omega$ and where $\mathbf{c}_2 = \Omega(\mu_3,\theta_2) - \mathbf{B}\mu_3$, we have that for any point $\mathbf{v}_i$ close to $\mu_3$ that $\Omega(\mathbf{v}_i,\theta_2) \approx \mathbf{B}\mathbf{v}_i + \mathbf{c}_2$. While computing $\{\mathbf{B},\mathbf{c}_2\}$ through linearization is generally very expensive, computing the approximation to \eqref{objective_all} does not require \emph{explicit} access to either $\mathbf{B}$ nor $\mathbf{c}_2$. Note that the second term in our proposed Eq \eqref{objective_all} for $l=1$ is given as 

\begin{align}
\label{mean_expression_simplified}
    & \ell\left(\mathbb{E}_{\mathbf{n}\sim \mathcal{N}(\mathbf{0},\Sigma_x)}[\Phi(\mathbf{x}_i + \mathbf{n};\theta)], y_i \right)  \notag\\
    =~ & \ell \left(\mathbb{E}_{\mathbf{z}_i\sim \mathcal{N}(\mathbf{x}_i,\Sigma_x)}[\Omega\left(\text{ReLU}_1 \left(\Upsilon(\mathbf{z}_i;\theta_1)\right);\theta_2\right)], y_i\right) \notag\\
    =~ & \ell \left(\mathbb{E}_{\mathbf{z}_i\sim \mathcal{N}(\mathbf{x}_i,\Sigma_x)}[\Omega\left( \text{ReLU}_1 \left(\mathbf{A}\mathbf{z}_i + \mathbf{c}_1\right);\theta_2\right)], y_i\right) \notag\\
    \approx~ & \ell \left(\mathbb{E}_{\mathbf{z}_i \sim \mathcal{N}(\mathbf{x}_i,\Sigma_x)}\left[\mathbf{B}\left(\text{ReLU}_1 \left(\mathbf{A}\mathbf{z}_i + \mathbf{c}_1\right)\right) + \mathbf{c}_2\right], y_i\right) \notag \\
    = ~ &  \ell \left(\mathbf{B}\mu_3 + \mathbf{c}_2, y_i\right) = \ell \left(\Omega(\mu_3,\theta_2), y_i\right). 
\end{align}

\noindent The approximation follows from the assumption that the input to the second subnetwork $\Omega(.;\theta_2)$, \ie $\mathbf{v}_i =  \text{ReLU}_1 \left(\mathbf{A}\mathbf{z}_i + \mathbf{c}_1\right))$, is close to the point of linearization $\mu_3$ such that $\Omega(\mathbf{v}_i;\theta_2) \approx \mathbf{B}\mathbf{v}_i + \mathbf{c}_2$. Or simply, that the input to $\Omega$ is close to the mean inputs, \ie $\mu_3$, to $\Omega$ under Gaussian noise. The penultimate equality follows from the linearity of the expectation. As for the last equality, note that $\{\mathbf{B},\mathbf{c}_2\}$ are the linearization parameters of $\Omega$ at $\mu_3$, where $\mathbf{c}_2 = \Omega(\mu_3,\theta_2) - \mathbf{B}\mu_3$ by the first order Taylor approximation. Thus, computing the second term of Eq \eqref{objective_all} according to Eq \eqref{mean_expression_simplified} can be simply approximated by a forward pass of $\mu_3$ through the second network $\Omega$. As for computing $\mu_3 = T(\mu_2,\sigma_2)$, note that $\mu_2 = \mathbf{A} \mathbf{x}_i + \mathbf{c}_1$ in Eq \eqref{mean_expression_simplified}, which is equivalent to a forward pass of $\mathbf{x}_i$ through the first subnetwork. This is since $\Upsilon(.,\theta_1)$ is linear, where $\theta_1 = \{\mathbf{A},\mathbf{c}_1\}$. Moreover, note that since $\sigma_2 = \sqrt{\text{diag}\left(\mathbf{A}\Sigma_x\mathbf{A}^\top\right)}$ where $\Sigma_x = \text{Diag}\left(\sigma_x^2\right)$, the following identity follows:

\begin{align}
\sigma_2 = \sqrt{\text{diag}\left(\mathbf{A} \text{Diag}\left(\sigma_x^2\right) \mathbf{A}^\top\right)} = \sqrt{\left(\mathbf{A} \odot \mathbf{A}\right) \sigma_x^2}.
\end{align}

\noindent The latter can be efficiently computed by simply squaring the linear parameters in the first subnetwork and performing a forward pass of the input noise variance $\sigma_x^2$ through $\Upsilon$ without the bias $\mathbf{c}_1$ and taking the element-wise square root. Lastly, it is straightforward to compute $T(\mu_2,\sigma_2)$ as it is an element-wise function as defined in Eq \eqref{theorem_1}. The overall computational graph in Figure \ref{fig:pipeline_fig} shows a summary on the computation of Eq \eqref{objective_all} using only forward passes through the two subnetworks $\Upsilon$ and $\Omega$. It is now possible with the proposed efficient approximation of our proposed regularizer in Eq \eqref{objective_all} to efficiently train networks on noisy training examples that are corrupted with noise $\mathcal{N}(\mathbf{0},\Sigma_x)$ without any form of prohibitive data augmentation.

\section{Experiments}
\label{experiments}
In this section, we conduct experiments on multiple network architectures and datasets to demonstrate the effectiveness of our proposed regularizer in training more robust networks, especially in comparison with data augmentation. We first propose a new unified robustness metric against additive noise from a general distribution $\mathcal{D}$. We later specialize this metric to the case when $\mathcal{D}$ is Gaussian. Lastly, we compare our experimental results to data augmentation.

\subsection{On the Robustness Evaluation Metric}
While there is a consensus on the definition of robustness in the presence of adversarial attacks, as the smallest perturbation required to fool a network, change the prediction of the network on the noise-free example, it is not straightforward to extend such a definition to additive noise sampled from a distribution $\mathcal{D}$. In particular, the work of \cite{Fawzi2018} tried to address this difficulty by defining the robustness of a classifier around an example $\mathbf{x}$ as the distance between $\mathbf{x}$ and the closest decision boundary. However, such a definition is difficult to compute in practice and is not scalable, as it requires solving a generally nonconvex optimization problem for every testing example $\mathbf{x}$ that may also suffer from poor local minima. To remedy these drawbacks, we present a new robustness metric for generic additive noise.

\noindent \paragraph{Robustness Against Additive Noise.} Consider a classifier $\Psi(.)$ with $\psi(\mathbf{x}) = \arg\max_i \Psi_i(\mathbf{x})$ as the predicted class label for the example $\mathbf{x}$ regardless of the correct class label $y_i$. We define the robustness on a sample $\mathbf{x}$ against a generic additive noise sampled from a distribution $\mathcal{D}$ as:

\begin{equation}
	\Re_{\mathcal{D}}(\mathbf{x}) = \mathbb{P}_{\mathbf{n}\sim\mathcal{D}}\{\psi(\mathbf{x}+\mathbf{n}) = \psi(\mathbf{x})\}.
	\label{eq:general_robustness}
\end{equation}

\noindent Here, the proposed robustness metric $\Re_{\mathcal{D}}(\mathbf{x})$ measures the probability of the classifier to preserve the original prediction of the noise-free example, \ie $\psi(\mathbf{x})$, after adding noise, \ie $\psi(\mathbf{x}+\mathbf{n})$, from distribution $\mathcal{D}$. Therefore, the robustness over a testing dataset $\mathcal{T}$ can be defined as the expected robustness over the test dataset: $\Re_{\mathcal{D}}(\mathcal{T}) = \mathbb{E}_{\mathbf{x} \sim \mathcal{T}}\left[\Re_{\mathcal{D}}(\mathbf{x})\right]$. Inspired by \cite{Jean2018}, for ease, we relax Definition \eqref{eq:general_robustness} from the probability of preserving the prediciton score under $\mathcal{D}$ sampled noise to a 0/1 robustness over $m$-randomly sampled examples from $\mathcal{D}$. That is we say that $\Re_{\mathcal{D}}(\mathbf{x}) = 1$, if for $m$ randomly sampled noise from $\mathcal{D}$ that are added to $\mathbf{x}$, none of which changed the prediction from $\psi(\mathbf{x})$. However, if a single example of the $m$ samples changed the prediction from $\psi(\mathbf{x})$, we set $\Re_{\mathcal{D}}(\mathbf{x}) = 0$. The final robustness score is then averaged over the testing dataset $\mathcal{T}$.

\noindent \paragraph{Robustness Against Gaussian Noise.} For additive Gaussian noise, \ie $\mathcal{D} = \mathcal{N}(\mathbf{0}, \Sigma_x = \text{Diag}\left(\sigma_x^2\right))$, robustness is averaged over a range of testing variances $\sigma_x^2$. We restrict $\sigma_x$ to 30 evenly sampled values in $[0,0.5]$, where this set is denoted as $\mathcal{A}$\footnote{We assume that the input $\mathbf{x}$ is normalized $[0,1]^n$}. In practice, this is equivalent to sampling $m$ Gaussian examples for each $\sigma_x \in \mathcal{A}$, and if none of the $m$ samples changes the prediction of the classifier $\psi$ from the original noise-free example, the robustness for that sample at that $\sigma_x$ noise level is set to $1$ and then averaged over the complete testing set. Then, the robustness is the average over multiple $\sigma_x \in \mathcal{A}$. To make the computation even more efficient, instead of sampling a large number of Gaussian noise samples ($m$), we only sample a single noise sample with the average energy over $\mathcal{D}$. That is we sample a single $\mathbf{n}$ of norm $\|\mathbf{n}\|_2 = \sigma_x \sqrt{n}$. This is due to the fact that:
\begin{align*}
    \underset{\mathbf{x}\sim\mathcal{N}\left(\mathbf{0}_n, \sigma_x^2\mathbf{I}\right)}{\mathbb{E}}\left[\|\mathbf{x}\|_2\right] = \sqrt{2\sigma_x^2}\frac{\Gamma\left(\frac{n+1}{2}\right)}{\Gamma\left(\frac{n}{2}\right)} \overset{n\to\infty}{=} {\sigma_x\sqrt{n}}    
\end{align*}

\begin{figure*}
	\centering
	\begin{subfigure}[b]{0.32\textwidth}
		\caption{MNIST}
		\centering \includegraphics[width=\textwidth]{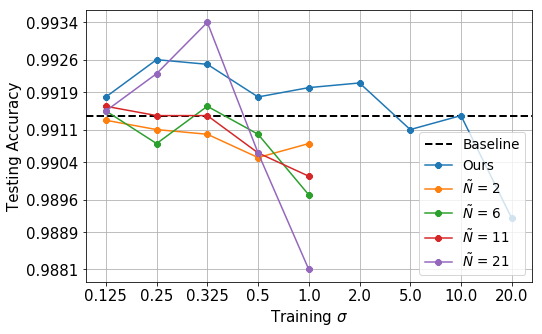}
	\end{subfigure}
	\hfill
	\begin{subfigure}[b]{0.32\textwidth}
		\caption{CIFAR10}
		\centering \includegraphics[width=\textwidth]{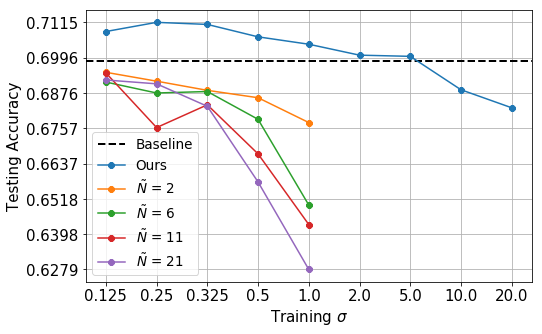}
	\end{subfigure}
	\hfill
	\begin{subfigure}[b]{0.32\textwidth}
		\caption{CIFAR100}
		\centering \includegraphics[width=\textwidth]{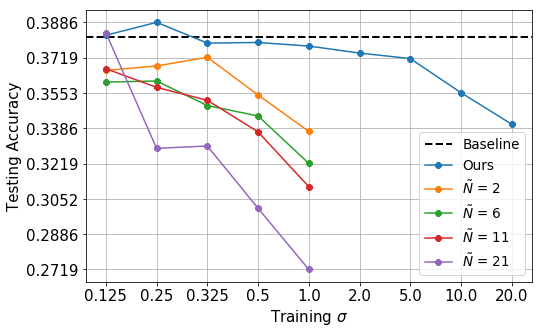}
	\end{subfigure}
	
	\vskip\baselineskip

	\begin{subfigure}[b]{0.32\textwidth}
		\centering \includegraphics[width=\textwidth]{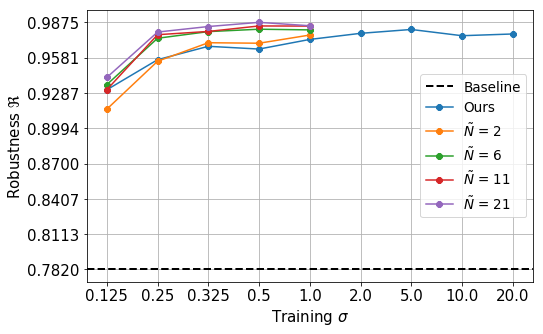}
	\end{subfigure}
	\hfill
	\begin{subfigure}[b]{0.32\textwidth}
		\centering \includegraphics[width=\textwidth]{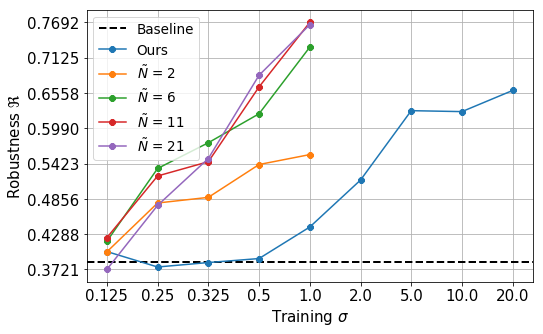}
	\end{subfigure}
	\hfill
	\begin{subfigure}[b]{0.32\textwidth}
		\centering \includegraphics[width=\textwidth]{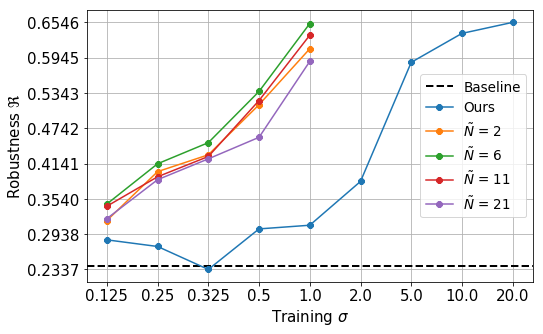}
	\end{subfigure}
	\caption{\textbf{General trade-off between accuracy and robustness on LeNet}. We see, in all plots, that the accuracy tends to be negatively correlated with robustness over varying noise levels and amount of augmentation. Baseline refers to training without data augmentation nor our regularizer. However, it is hard to compare the performance of our method against data augmentation from these plots as we can only compare the robustness of models with similar noise-free testing accuracy.}
	\label{fig:lenet_trend}
\end{figure*}

\subsection{Experimental Setup}
In this section, we demonstrate the effectiveness of the proposed regularizer in improving robustness. Several experiments are performed with our objective \eqref{objective_all}, where we strike a comparison with data augmentation approaches. 

\noindent \paragraph{Hyperparameters.} We opt to use PyTorch version 0.4.1 \cite{paszke2017pytorch} to implement all the experiments\footnote{Source code: \href{https://github.com/ModarTensai/gaussian-regularizer}{github.com/ModarTensai/gaussian-regularizer}}, with the hyper-parameters fixed as listed in Table \ref{tbl:parameters} and two $\mathrm{optimizer}$s: \mbox{Adam (betas=$(0.9, 0.999)$, eps=${10}^{-8}$, amsgrad=False) \cite{kingma2014adam}}, \mbox{SGD (momentum=$0.9$, dampening=$0$, nesterov=True) \cite{loshchilov2016sgdr}}. In each experiment, we randomly split the training dataset into 10\% validation and 90\% training and monitor the validation loss after each epoch. If it didn't improve for $\mathrm{lr\_patience}$ epochs, we reduce the learning rate by multiplying it by $\mathrm{lr\_factor}$ and we start with an initial learning rate of $\mathrm{lr\_initial}$. The training is terminated only if the validation loss didn't improve for $\mathrm{loss\_patience}$ epochs or if training reached $100$ epochs, and report the results on the model with the best validation loss.

\noindent \paragraph{Architecture Details.} The input images in MNIST (gray-scale) and CIFAR (colored) are squares with sides equal to $28$ and $32$, respectively. Since AlexNet was originally trained on ImageNet of sides equal to $224$, we will marginally alter the implementation of AlexNet in TorchVision \cite{marcel2010torchvision} to accommodate for this difference. First, we need the output of the fully-convolutional part of the models to have a positive shape, which lead to changing the number of hidden units in the first fully-connected layer (in LeNet to $4096$, AlexNet to $256$, LeNet on MNIST to $3136$). For AlexNet, we will changed all pooling kernel sizes from $3$ to $2$ and the padding size of $\mathrm{conv1}$ from $2$ to $5$. Second, we swapped each $\mathrm{maxpool}$ with the preceding ReLU which will make training and inference more efficient. Third, we needed to enforce that the first layer in all the models is a convolution followed by ReLU as discussed earlier. Lastly, for analysis simplicity, we removed all $\mathrm{dropout}$ layers.

\begin{table}
	\centering
	\begin{tabular}{l|l|l|}
	\hline
		Hyper-parameter            & LeNet  & AlexNet  \\ \hline\hline 
		$\mathrm{optimizer}$       & Adam   & SGD      \\ \cline{2-3} 
		$\mathrm{minibatch\_size}$ & 1000   & 128      \\ \cline{2-3} 
		$\mathrm{lr\_initial}$     & 0.0001 & 0.1      \\ \cline{2-3} 
		$\mathrm{lr\_patience}$    & 3      & 5        \\ \cline{2-3} 
		$\mathrm{lr\_factor}$      & 0.9    & 0.5      \\ \cline{2-3} 
		$\mathrm{loss\_patience}$  & 10     & 20       \\ \cline{2-3} 
		$\mathrm{weight\_decay}$   & 0      & 0.0005   \\ \cline{2-3} 
		\hline
		\hline
	\end{tabular}
	\caption{Lists the training optimization hyper-parameters.}
	\label{tbl:parameters}
\end{table}

\subsection{Results}
For each model and dataset, we compare vanilla trained models, models trained with noise-free data and without our regularization, against two groups of experiments; data augmentation and our objective \eqref{objective_all}. Each group has two configurable variables: the level of noise controlled by $\sigma_x^2$ during training, and the amount of noise controlled by the trade-off coefficient $\alpha$ in Eq \eqref{objective_all} or $\tilde{N}$ in the case of augmentation, where $\tilde{N}$ is the number of added noisy training examples.

\begin{figure*}
	\centering
	\begin{subfigure}[b]{0.32\textwidth}
	    \caption{MNIST (acc = $99.14\%$, $\tilde{N}$ = $21$)}
		\centering \includegraphics[width=\textwidth]{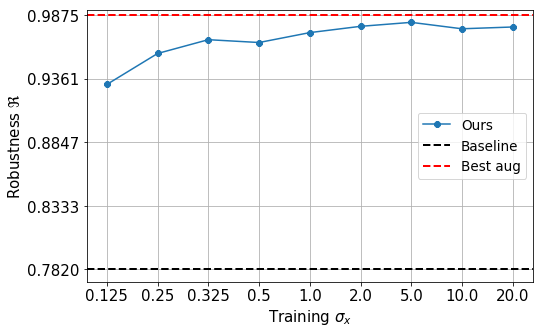}
		\label{fig:lenet_mnist}
	\end{subfigure}
	\hfill
	\begin{subfigure}[b]{0.32\textwidth}
	    \caption{CIFAR10 (acc = $69.85\%$, $\tilde{N}$ = $2$)}
		\centering \includegraphics[width=\textwidth]{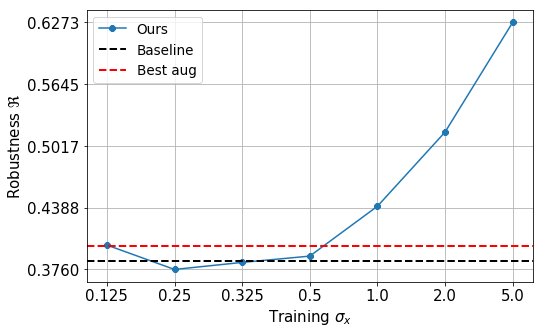}
		\label{fig:lenet_cifar10}
	\end{subfigure}
	\hfill
	\begin{subfigure}[b]{0.32\textwidth}
	    \caption{CIFAR100 (acc = $38.16\%$, $\tilde{N}$ = $21$)}
		\centering \includegraphics[width=\textwidth]{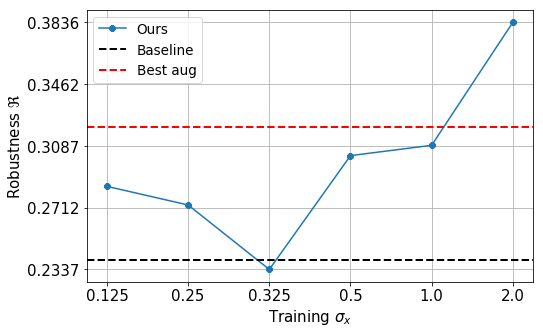}
		\label{fig:lenet_cifar100}
	\end{subfigure}
	\vspace{-0.5cm}
	\caption{\textbf{Fair robustness comparison of LeNet with data augmentation and our regularizer}. The results reported are only for models with a testing accuracy that is at least as good as the accuracy of the baseline model with a tolerance; $0\%$, $0.39\%$, and $0.75\%$ for MNIST, CIFAR10, CIFAR100, respectively. Thereafter, only the models with the highest robustness are presented. It is clear that training with our regularizer, while maintaining a high noise-free testing accuracy, can attain similar/better robustness than performing 21 times noisy data augmentation on MNIST and CIFAR100.}
	\label{fig:lenet_robustness}
	\vspace{-0.5cm}
\end{figure*}

\begin{figure*}
	\centering
	\begin{subfigure}[b]{0.475\textwidth}
	    \caption{CIFAR 10 (acc = $68.91\%$, $\tilde{N}$ = $11$)}
		\centering \includegraphics[width=\textwidth]{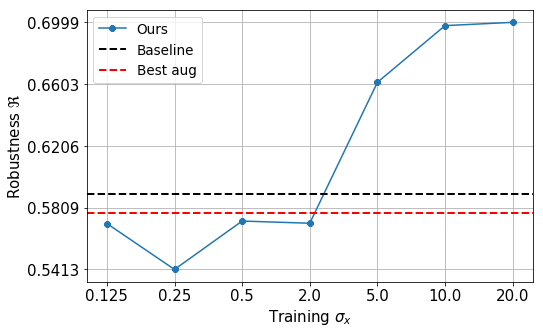}
	\end{subfigure}
	\begin{subfigure}[b]{0.475\textwidth}
	    \caption{CIFAR 100 (acc = $38.10\%$, $\tilde{N}$ = $6$)}
		\centering \includegraphics[width=\textwidth]{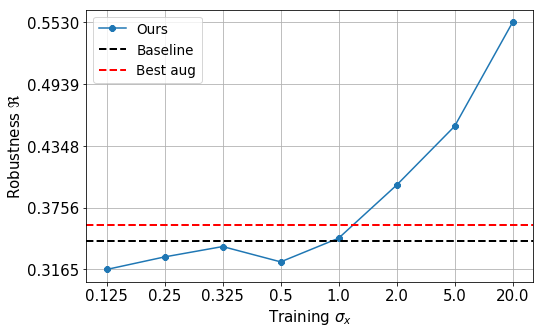}
	\end{subfigure}
	\vspace{-0.25cm}
	\caption{\textbf{Fair robustness comparison of AlexNet with data augmentation and our regularizer}. The reported models trained with our regularizer on CIFAR10 and CIFAR100 on all training $\sigma_x$ are within $1.68\%$ and $4.83\%$ accuracy of the baseline accuracy, respectively. Note that the models trained with the proposed regularizer can achieve better robustness than 11-fold and 6-fold noisy data augmentation on CIFAR10 and CIFAR100, respectively.}
	\label{fig:alexnet_robustness}
	\vspace{-0.15cm}
\end{figure*}

\noindent\paragraph{Accuracy vs. Robustness.} We start by demonstrating that data augmentation tend to improve the robustness, as captured by $\Re(\mathcal{T})$ over  the testset, on the expense of decreasing the testing accuracy on the noise-free examples. Realizing this is essential for a fair comparison, as one would need to compare the robustness of networks that only have similar noise-free testing accuracies. To show this, we ran $60$ training experiments with data augmentation on LeNet with three datasets (MNIST, CIFAR10, and CIFAR100), four augmentation levels ($\tilde{N} \in \{2, 6, 11, 21\}$), and five noise levels ($\sigma_x \in \mathcal{A} = \{0.125, 0.25, 0.325, 0.5, 1.0\}$). In contrast, we ran robust training experiments using our objective \eqref{objective_all} with the trade-off coefficient $\alpha \in \{0.5, 1, 1.5, 2, 5, 10, 20\}$ on the same datasets, but we extended the noise levels $\sigma_x$ to include the extreme noise regime of $\sigma_x\in\{2, 5, 10, 20\}$. These noise levels are too large to be used for data augmentation, especially since $\mathbf{x} \in [0,1]^n$; however as we will see, they are still beneficial for our proposed regularizer. Figure \ref{fig:lenet_trend} shows both the testing accuracy and robustness as measured by $\Re(\mathcal{T})$ over a varying range of training $\sigma_x$ for the data augmentation approach (with varying augmentation levels) of LeNet on MNIST, CIFAR-10 and CIFAR-100. It is important to note here that the main goal of these plots is not to compare the robustness achieved by our method versus data augmentation, rather, they demonstrate a very important trend. In particular, increasing the training $\sigma_x$ for either approach degrades testing accuracy on noise-free data. However, the degradation in our approach is much more graceful since the trained LeNet model was never directly exposed to individually corrupted examples during training as opposed to the data augmentation approach. Note that our proposed regularizer \eqref{objective_all} enforces the separation between the expected output prediction analytically. Moreover, the robustness of both methods consistently improves as the training $\sigma_x$ increases. This trend holds even on the easiest dataset (MNIST). Interestingly, models trained with our regularizer enjoys an improvement int testing accuracy over the baseline vanilla trained model. Such a behaviour is only exhibited with large factor of augmentation, $\tilde{N}=21$, with a small enough training $\sigma_x$ on MNIST. This indicates that models can benefit from better accuracy with a good approximation of \eqref{objective_all_or} through our proposed objective \eqref{objective_all} or through extensive Monte Carlo estimation. However, as $\sigma_x$ increase, Monte Carlo estimates of the second term in \eqref{objective_all_or} through data augmentation, with $\tilde{N}=21$, is no longer enough to capture the noise. In the next section, we properly compare the robustness among networks that have a similarly high testing accuracy for a fair comparison.

\noindent\paragraph{Robustness Comparison.} Following the previous discussion, it is essential to only compare the robustness of networks that achieve similar testing accuracy as it would be unfair otherwise. This is since one can achieve a perfect robustness with a deterministic classifier that assigns the same class label to all possible inputs. This is mainly due to the proposed robustness metric, as most common robustness metrics, being disassociated from the ground-truth labels where only the predictions of the model are considered. Therefore, we filtered out the results from Figure \ref{fig:lenet_trend} by removing all the experiments that achieved lower accuracy than the baseline model with some tolerance while comparing against the best robustness achieved through data augmentation in that range of accuracy and the baseline. Figure \ref{fig:lenet_robustness} summarizes the results for LeNet. Now, we can clearly see the difference between training with data augmentation and our approach. For MNIST (Figure \ref{fig:lenet_mnist}), we achieved the same robustness as 21-fold data augmentation without feeding the network with any noisy examples during training and while preserving the same baseline accuracy. Better yet, for CIFAR10 (Figure \ref{fig:lenet_cifar10}), our method is twice as robust as the best robustness achieved via data augmentation. Moreover, for CIFAR100 (Figure \ref{fig:lenet_cifar100}), we are able to outperform data augmentation by around $5\%$. Finally, for extra validation, we also executed the same experiments with a different model (AlexNet) on CIFAR10 and CIFAR100. In Figure \ref{fig:alexnet_robustness}, we present the same results with similar observations and conclusions. We can see that our proposed regularizer can improve robustness by $15\%$ on CIFAR10 and around $25\%$ on CIFAR100. It is interesting to note that for CIFAR10, data augmentation could not improve the robustness of the trained models without drastically degrading the testing accuracy on the noise-free examples. Moreover, it is interesting to observe that the best robustness achieved through data augmentation is even worse than the baseline. This could be due to the trade-off coefficient $\alpha$ in \eqref{objective_all_or}.

\section{Conclusion}
\label{conclusion}

Addressing the sensitivity problem of deep neural networks to adversarial perturbation is of great importance to the vision community. However, building robust classifiers against these noises is computationally expensive as it is generally done through the means of data augmentation. We proposed a generic lightweight analytic regularizer, which can be applied to any deep neural network with a ReLU activation after the first affine layer, that is designed to increase the robustness of the trained models under additive Gaussian noise. We demonstrated this with multiple architectures and datasets and showed that it outperforms data augmentation without observing any noisy examples.

\noindent \paragraph{Acknowledgments.} This work was supported by the King Abdullah University of Science and Technology (KAUST) Office of Sponsored Research.

{\small
\bibliographystyle{ieee}
\bibliography{main}
}

\end{document}